\documentclass[10pt,letterpaper]{article}

\usepackage[margin=1.9cm]{geometry}
\usepackage[none]{hyphenat}
\usepackage{hyperref}
\usepackage{graphicx}
\usepackage{caption}
\usepackage{fvextra}

\begin{document}

\title{Framing AI System Benchmarking as a Learning Task:\\
FlexBench and the Open MLPerf Dataset}

\author{
  Grigori Fursin\\
  Head of FCS Labs, FlexAI \\ 
  Founder, cTuning (AI Systems R\&D) \\ 
  \and
  Daniel Altunay\\
  Machine Learning Engineer, FlexAI\\
}

\maketitle

\begin{center}
  \href{https://github.com/mlcommons/ck/tree/master/flexbench}{github.com/mlcommons/ck/tree/master/flexbench}\\
  \href{https://cKnowledge.org}{cKnowledge.org}
\end{center}

\vskip 0.2in

\begin{abstract}

Existing AI system benchmarks such as MLPerf often struggle to keep pace
with the rapidly evolving AI landscape, making it difficult to
support informed deployment, optimization, and co-design decisions for AI
systems.
We suggest that benchmarking itself can be framed as an AI task—one in
which models are continuously evaluated and optimized across diverse
datasets, software, and hardware, using key metrics such as accuracy,
latency, throughput, energy consumption, and cost.
To support this perspective, we present \textbf{FlexBench}: a modular
extension of the MLPerf LLM inference benchmark, integrated with Hugging
Face and designed to provide relevant and actionable insights. 
Benchmarking results and metadata are collected into an \textbf{Open MLPerf Dataset}, 
which can be collaboratively curated, extended, and leveraged for \textbf{predictive
modeling and feature engineering}. 
We successfully validated the FlexBench concept through \textbf{MLPerf
Inference submissions}, including evaluations of \textbf{DeepSeek R1} and
\textbf{LLaMA 3.3} on commodity servers.
The broader objective is to enable practitioners to
make cost-effective AI deployment decisions that reflect their available
resources, requirements, and constraints.

\end{abstract}

\vspace{0.3cm}


\vspace{0.3cm}

\section{Motivation}

AI service providers, server manufacturers, and data center operators face
a critical challenge of selecting the right hardware and software configurations 
that achieve a return on investment within 3–5 years, despite a rapidly
changing AI ecosystem~\cite{mad-landscape}. MLPerf was introduced as a
full-stack inference benchmark to provide standardized, reproducible, and
comparable evaluations of accuracy, latency, and throughput across diverse
hardware and software stacks~\cite{reddi2019mlperf}.

Traditional benchmarks, however, are inherently limited. The number of
possible combinations of models, datasets, algorithms, and hardware is
immense, making exhaustive evaluation both impractical and costly. For
example, Hugging Face already hosts over a million models, tens of
thousands of datasets, and many methods, while new hardware and
software stacks continue to emerge. Comprehensive benchmarking under these
conditions remains one of the most significant open problems.

Moreover, MLPerf benchmarks cover only a small set of combinations—
typically a dozen or fewer—and are updated only annually, making it
difficult to capture the pace of AI research and deployment. Current MLPerf LLM
benchmarks still emphasize models such as BERT, LLaMA 2, and LLaMA
3, while newer systems including LLama 4, GPT-5, DeepSeek and Gemma 
are already widely adopted.

Furthermore, our extensive hands-on experience with MLPerf shows that the
heavily over-optimized results reported by a few chip manufacturers are
rarely reproducible out of the box on other models, software versions,
or hardware configurations—greatly limiting their practical usefulness.

\section*{FlexBench: An Alternative Approach}

Building on our previous work in applying AI techniques to systems research,
we propose treating benchmarking itself as a learning problem~\cite{29db2248aba45e59:a31e374796869125,acm_techtalk_fursin_reproducibility_2021,acm_rep_23_cm_keynote,fursin2024enablingefficientcosteffectiveaiml}.
Specifically, MLPerf benchmarking and AI system co-design can be
reformulated as a learning task supported by an open dataset of results
and trainable objective functions that optimize key metrics such
as accuracy, latency, throughput, power consumption, and cost.

\begin{figure*}[h]
  \centering
  \includegraphics[width=1\textwidth]{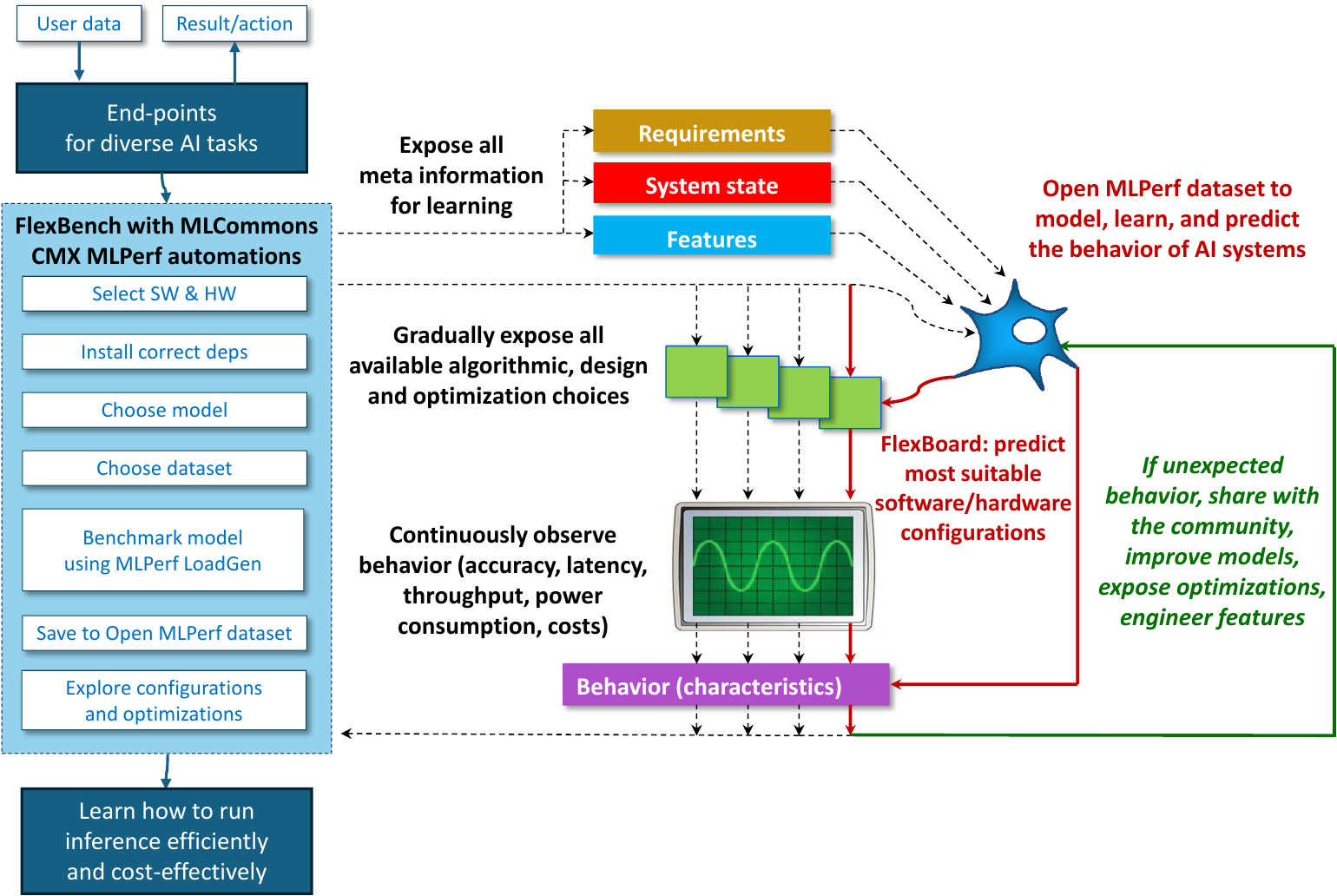}
  \caption{High-level architecture of FlexBench.}
  \label{fig:architecture}
\end{figure*}

To support this vision, we introduce \textbf{FlexBench}, an
open-source, modular amd flexible benchmarking framework derived from the MLPerf LLM
inference benchmark and connected with the Hugging Face Hub~\cite{flexbench-and-flexboard-github}. 
With a unified CLI and codebase, users can automatically benchmark a wide variety 
of models and datasets by modifying only a few input parameters. Supported
by the MLCommons CMX workflow automation framework with MLPerf automations~\cite{github_ck_cmind_cm_cmx,acm_rep_23_cm_keynote,fursin2024enablingefficientcosteffectiveaiml}, 
FlexBench is designed to evolve continuously alongside the AI ecosystem, as illustrated in Figure~\ref{fig:architecture}.

Results generated with FlexBench and standard MLPerf are aggregated via
MLCommons CMX into the \textbf{Open MLPerf Dataset}. This dataset
is openly shared on GitHub and Hugging
Face~\cite{open-mlperf-dataset-hf,flexbench-and-flexboard-github}, where
it can be collaboratively cleaned, extended, and analyzed using standard
data analytics techniques, including predictive modeling and feature
engineering. \textbf{FlexBoard} is then used to visualize, compare, and
predict the most efficient and cost-effective software/hardware
configurations for different models based on user requirements and
constraints.

\section*{Technical Overview}

\begin{figure*}[h]
  \centering
  \includegraphics[width=0.8\textwidth]{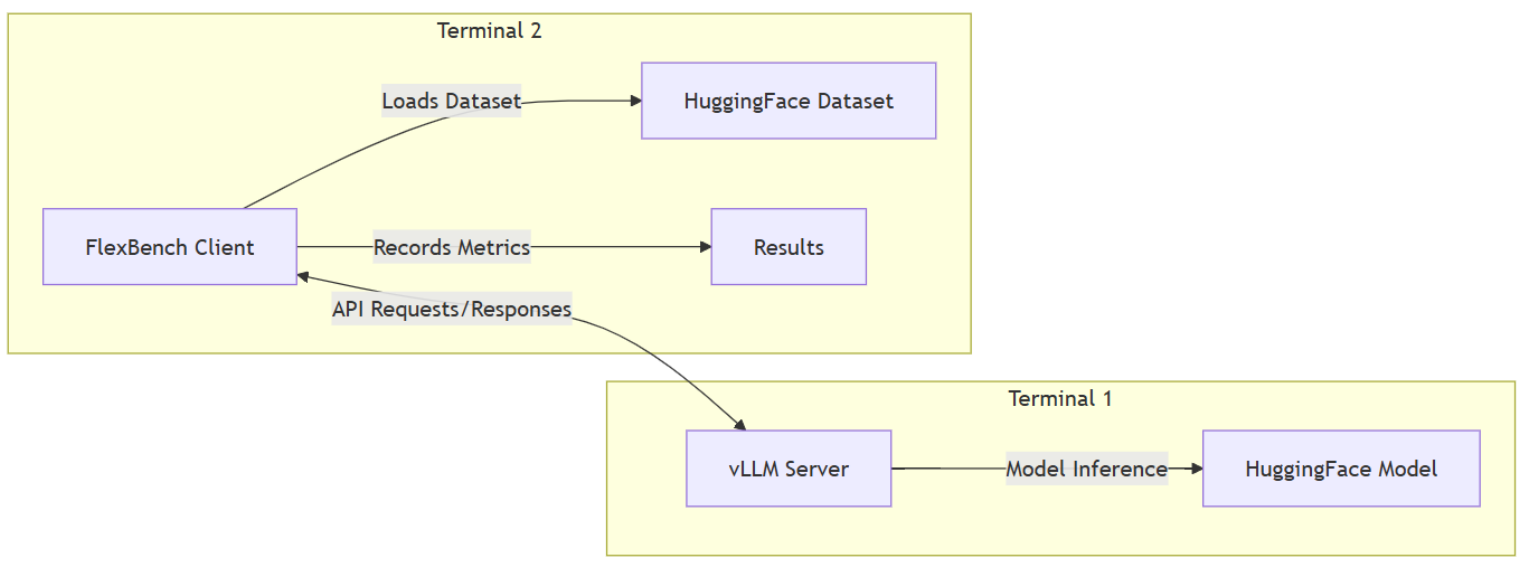}
  \caption{FlexBench client–server architecture.}
  \label{fig:client-server}
\end{figure*}

FlexBench employs a client–server design (Figure~\ref{fig:client-server}),
where the client connects to a running vLLM
server~\cite{vllm-github,kwon2023efficient}. It is built on MLPerf
LoadGen, the official MLPerf harness for measuring inference
performance~\cite{mlperf-loadgen-github,reddi2019mlperf}. By abstracting
models and datasets as interchangeable modules, FlexBench preserves
MLPerf’s rigor while enabling greater flexibility. Both Hugging Face and
local LLMs and datasets can be used with minimal setup.

\begin{figure*}[h]
  \centering
  \includegraphics[width=0.5\textwidth]{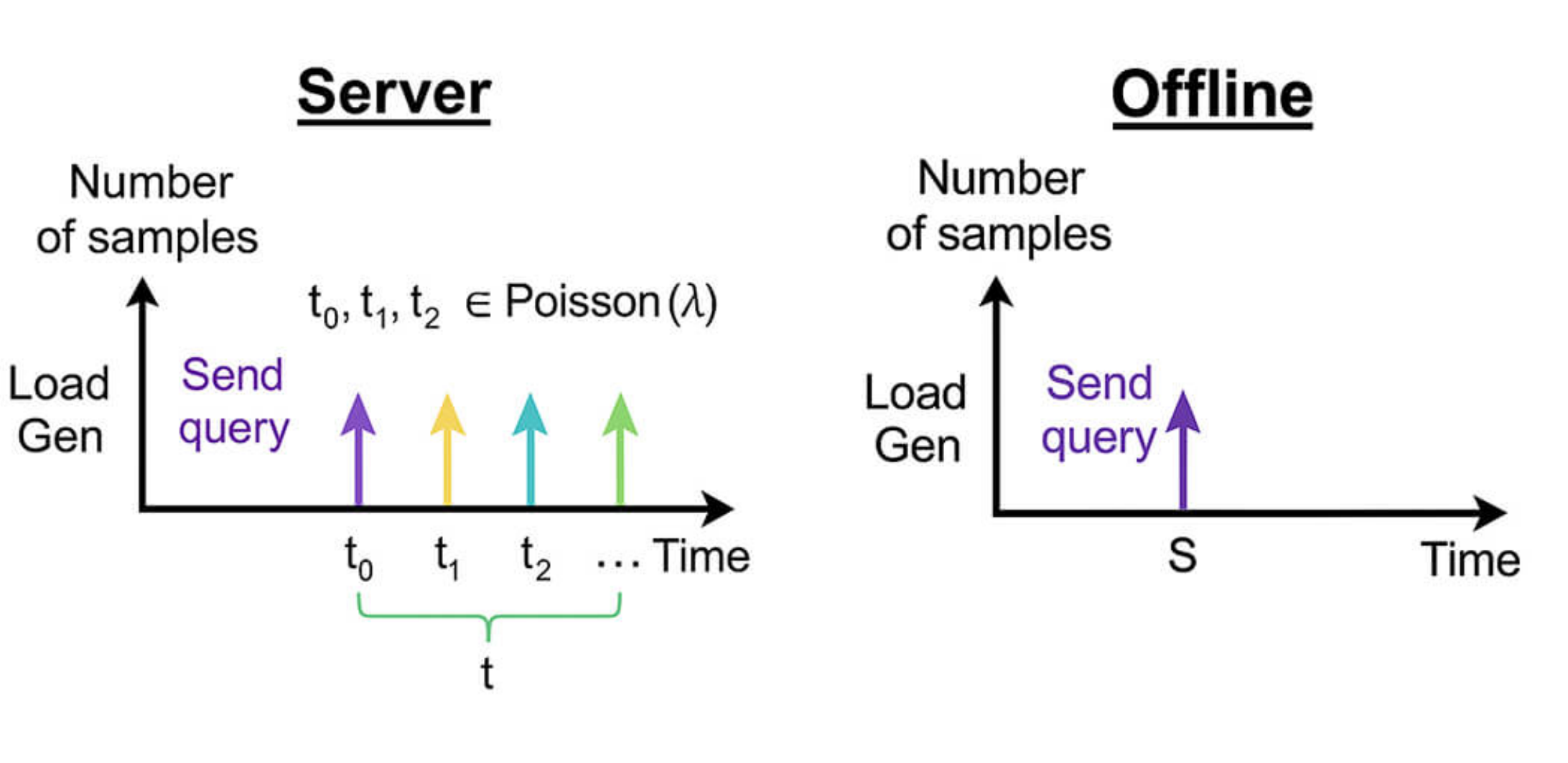}
  \caption{MLPerf LoadGen server and offline modes.}
  \label{fig:loadgen-server-offline}
\end{figure*}

Our framework supports the two standard inference modes defined in the
MLPerf Inference paper: \emph{Server} (streaming requests) and
\emph{Offline} (batch queries)~\cite{reddi2019mlperf}, as illustrated
in Figure~\ref{fig:loadgen-server-offline}. It reports detailed LoadGen
metrics—including throughput, latency distributions, and
time-to-first-token (TTFT)—that are fully compatible with MLPerf standards
and suitable for dataset inclusion. Moreover, FlexBench results have been
cross-validated against the native vLLM benchmarking
infrastructure~\cite{vllm-benchmark-github}, showing only minimal
discrepancies while also providing additional metrics, such as accuracy,
that are essential for further pruning, quantization, and fine-tuning
of models.

FlexBoard, implemented as a Gradio application, loads the Open MLPerf
dataset generated by MLCommons CMX automations~\cite{flexboard-hf}.
It provides predictive modeling and visualization capabilities to support
the selection of suitable software and hardware configurations for
different models under user-specified constraints. This functionality
builds on prior work in cost-efficiency analysis across heterogeneous
accelerators~\cite{ck-cost-efficiency-dashboard}.

\section*{Preliminary Results}

We validated our approach in the MLPerf Inference 5.0 submission by
benchmarking several non-MLPerf LLMs (e.g., DeepSeek R1 Distill LLaMA 8B, 
LLaMA 3.3) on the OpenOrca dataset using commodity servers with NVIDIA H100 GPUs. Our
automation framework enabled rapid switching among models, datasets, and
hardware configurations by adjusting command-line arguments, without
requiring code modifications.

We curated the Open MLPerf dataset by integrating past MLPerf results with
new FlexBench outputs. To enable predictive modeling, we cleaned the
dataset, standardized heterogeneous fields, and engineered additional
features such as model size and data type. A sample entry is shown below:

\begin{Verbatim}[fontsize=\footnotesize,breaklines=true]
{
  "metrics.accuracy": "ROUGE1: 30.6202  ROUGE2: 13.9221  ROUGEL: 18.9101 TOKENS_PER_SAMPLE: 581.8",
  "metrics.result": 2631.93,   "metrics.result\_per\_accelerator": 2631.93,
  "metrics.units": "Tokens/s",    "model.architecture": "LLM",
  "model.mlperf\_name": "llama2-70b-99",
  "model.name": "DeepSeek-R1-Distill-Llama-8B",
  "model.number\_of\_parameters": 8.0,
  "model.weight\_data\_types": "bfloat16",
  "software.framework": "vLLM v0.7.3",
  "software.operating\_system": "Ubuntu 22.04.5 LTS (5.15.0-131-generic)",
  "submission.availability": "available",   "submission.division": "open",
  "submission.organization": "FlexAI",   "submission.scenario": "Server",
  "system.accelerator.count\_per\_node": 1,
  "system.accelerator.name": "NVIDIA H100 80GB HBM3",
  "system.accelerator.total\_count": 1,
  "system.accelerator.vendor": "NVIDIA",
  "system.cpu.caches": "L1d cache: 6.3 MiB (200 instances), L1i cache: 6.3 MiB (200 instances), L2 cache: 800 MiB (200 instances), L3 cache: 3.1 GiB (200 instances)",
  "system.cpu.core\_count": 52,
  "system.cpu.count\_per\_node": 2,
  "system.cpu.model": "Intel Xeon Processor (SapphireRapids)",
  "system.interconnect.accelerator": "NVLink",  "system.interconnect.accelerator\_host": "PCIe",
  "system.name": "flexbench test node 0ef307db09d34a91 with 8xH100",
  "system.number\_of\_nodes": 1, "system.type": "datacenter"
}
\end{Verbatim}

\begin{figure*}[h]
  \centering
  \includegraphics[width=0.65\textwidth]{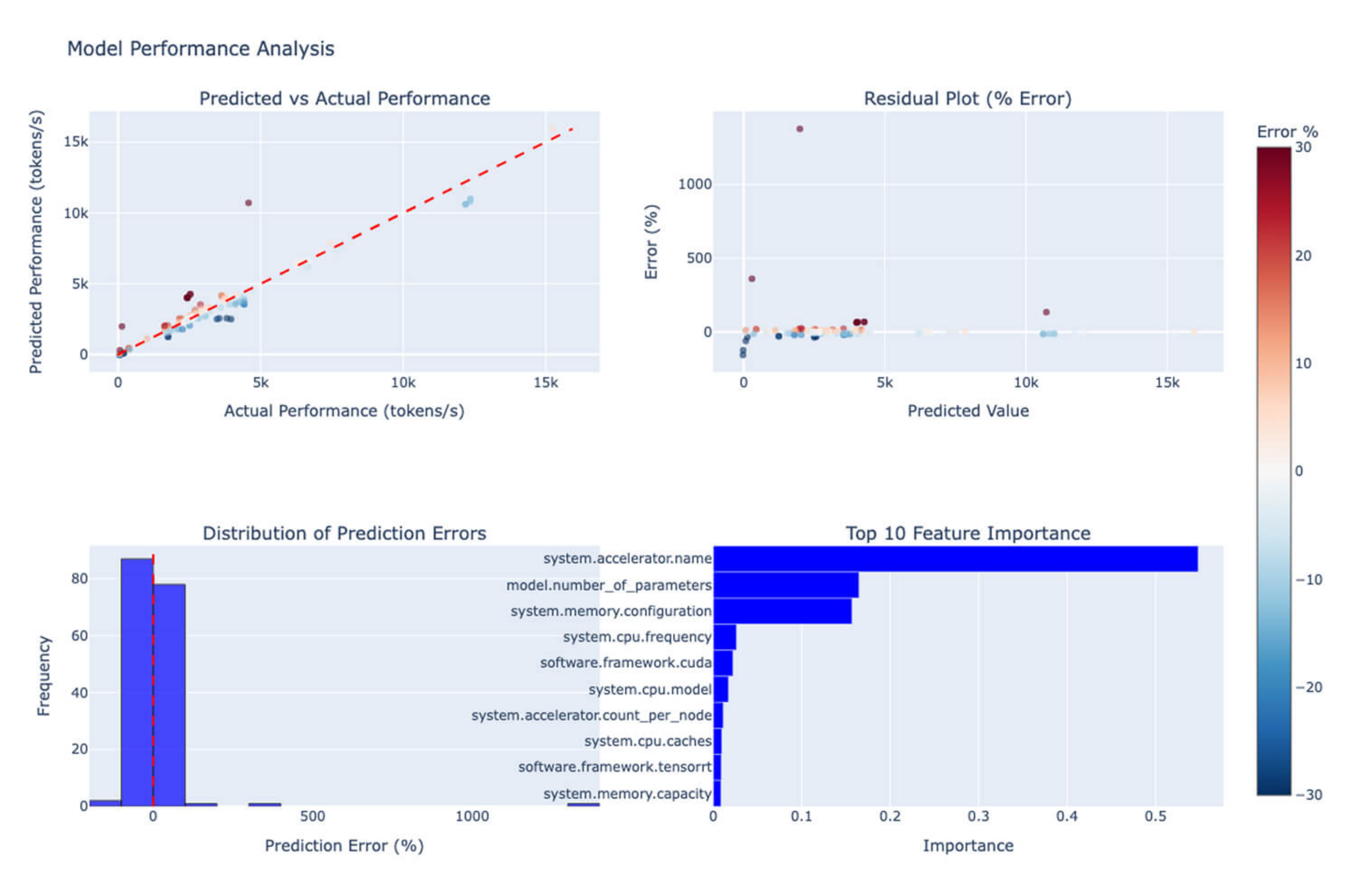}
  \includegraphics[width=0.4\textwidth]{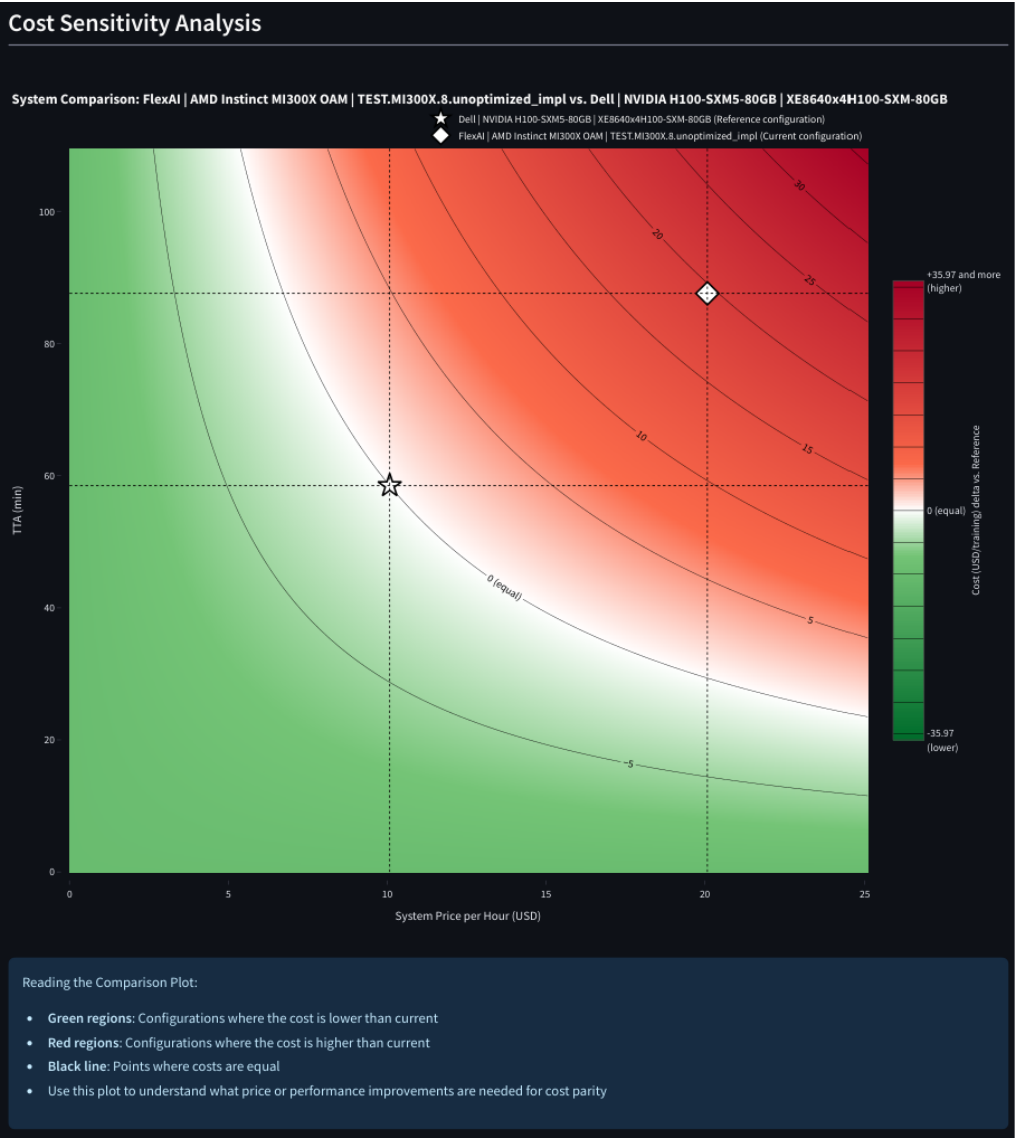}
  \includegraphics[width=0.5\textwidth]{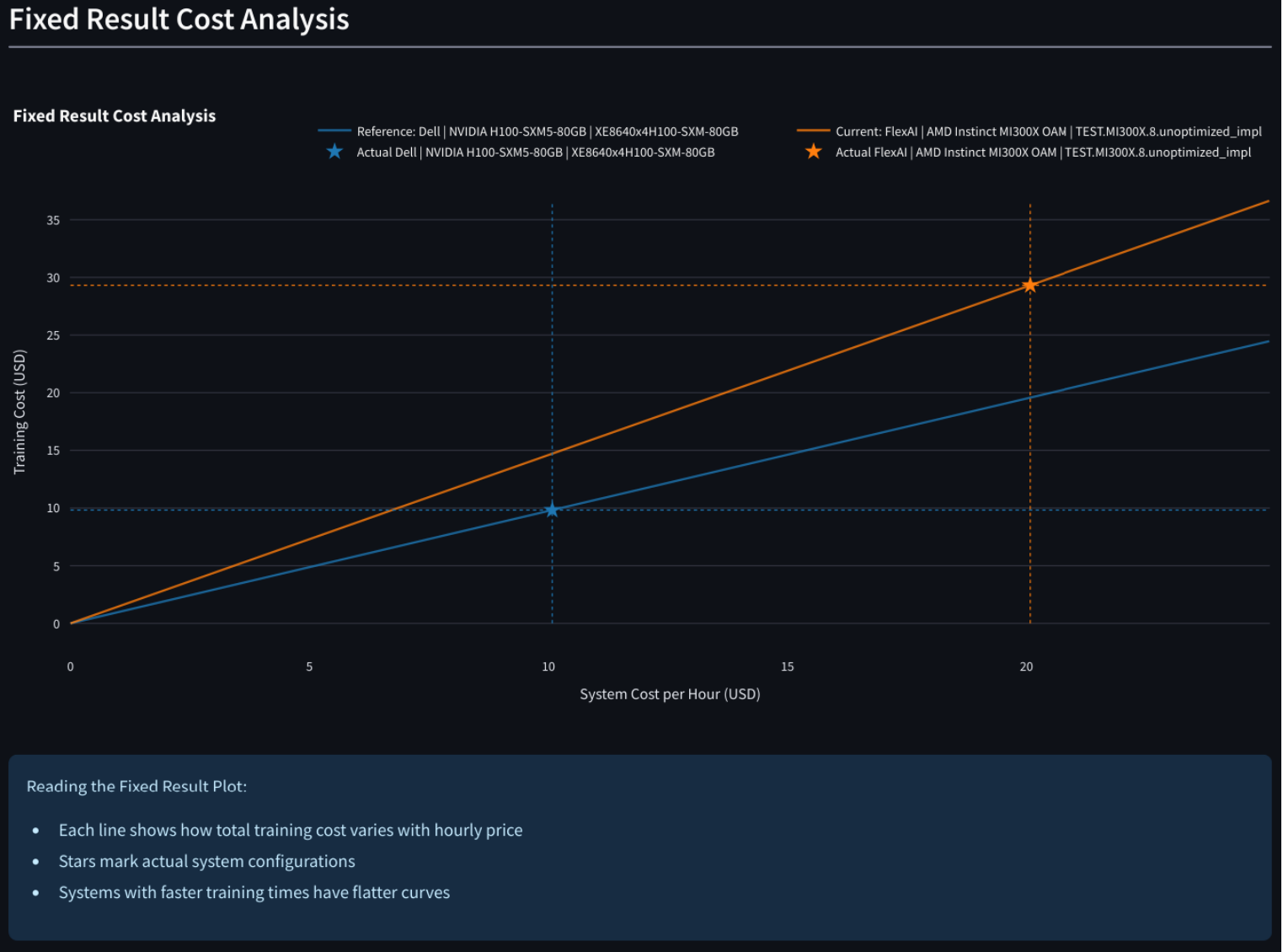}
  \caption{Examples of performance and cost analyses using FlexBench and FlexBoard.}
  \label{fig:dashboard_examples}
\end{figure*}

The cleaned dataset, FlexBoard visualization tool, and predictive modeling
utilities are openly released under Apache 2.0 license to assist researchers and practitioners
in improving benchmarking, evaluation, and optimization efforts.
For example, our proof-of-concept prototype allows beta users 
to input system costs and predict optimal software and hardware configurations 
based on model size and data type features (Figure~\ref{fig:dashboard_examples}).

\section*{Future Work}

FlexBench, FlexBoard, and CMX are still in the early stages of prototyping~\cite{flexbench-and-flexboard-github}. 
Planned directions include:

\begin{itemize}
\item Supporting additional models, datasets, and system configurations
\item Expanding the Open MLPerf dataset with the latest MLPerf results and new FlexBench outputs
\item Incorporating richer features (e.g., model graphs, tensor shapes, compiler optimizations, accelerator characteristics) 
      to improve predictions of optimal software/hardware configurations for previously unseen AI workloads
\item Extending FlexBoard functionality in response to user requirements
\item Engaging with the community to address practical benchmarking needs
\item Integrating FlexBench, FlexBoard and Open MLPerf dataset with the Collective Knowledge ecosystem~\cite{access-cknowledge}
\end{itemize}

Our long-term goal is to enable anyone to run AI models efficiently 
and cost-effectively, tailored to their available resources, requirements, and constraints. 
We also aim to help hardware manufacturers co-design more energy-efficient
AI systems and data centers, thereby reducing total cost of ownership and
accelerating time to market.

\section*{Acknowledgements}

We thank Dali Kilani, Salam Almosawi, Venkataraju Koppada, Rahul
Thangallapally, Aymen Zayet, Quentin Casasnovas, Agustin Mautone, Matthieu
Paret, Pierre-Yves Laligand, Jean-Noel Quintin, Jean-Baptiste Louazel,
Diego Coy, Harshal Patil, Alex Denisov, and others for their valuable
discussions and feedback. We also acknowledge the original FlexAI blog
post in which we first introduced FlexBench and the Open MLPerf
Dataset~\cite{flexai_blog}.

\bibliographystyle{plain}
\bibliography{flexbench}

\end{document}